\documentclass[manuscript,screen,nonacm]{acmart}
\usepackage{booktabs}
\usepackage{tabularx}
\usepackage{array}
\usepackage{makecell}
\usepackage{ragged2e}
\usepackage{seqsplit}
\newcolumntype{L}[1]{>{\RaggedRight\arraybackslash}p{#1}}

\AtBeginDocument{%
  }

\begin{document}

\title{Measuring Inclusion in Interaction: Inclusion Analytics for Human-AI Collaborative Learning}

\author{Jaeyoon Choi}
\email{jaeyoon.choi@uci.edu}
\affiliation{%
  \institution{University of California, Irvine}
  \city{Irvine}
  \state{California}
  \country{USA}
}
\author{Nia Nixon}
\email{dowelln@uci.edu}
\affiliation{%
  \institution{University of California, Irvine}
  \city{Irvine}
  \state{California}
  \country{USA}
}
\renewcommand{\shortauthors}{Choi and Nixon}

\begin{abstract}
Inclusion, equity, and access are widely valued in AIED, yet are often assessed through coarse sample descriptors or post-hoc self-reports that miss how inclusion is shaped moment by moment in collaborative problem solving (CPS). In this proof-of-concept paper, we introduce \textit{inclusion analytics}, a discourse-based framework for examining inclusion as a dynamic, interactional process in CPS. We conceptualize inclusion along three complementary dimensions ---participation equity, affective climate, and epistemic equity---and demonstrate how these constructs can be made analytically visible using scalable, interaction-level measures. Using both simulated conversations and empirical data from human-AI teaming experiments, we illustrate how inclusion analytics can surface patterns of participation, relational dynamics, and idea uptake that remain invisible to aggregate or post-hoc evaluations. This work represents an initial step toward process-oriented approaches to measuring inclusion in human-AI collaborative learning environments.
\end{abstract}





\maketitle

\section{Introduction}
The AIED community has long articulated inclusion, equity, and access as core values guiding the design and evaluation of intelligent learning technologies \cite{roscoe2022inclusion}. These commitments are reflected in research agendas, society mission statements, recurring conference themes, equity-focused workshops, and community initiatives and awards. Collectively, these signals emphasize a broad consensus that learning technologies should support diverse learners and reduce inequities rather than amplify them. However, while inclusion is widely valued, it remains far less clear how it should be understood, operationalized, and measured, particularly within the dynamics of learning activity itself\textbf{.}

In practice, inclusion has often been operationalized through relatively coarse indicators, such as representation counts, demographic breakdowns of study samples, or aggregate pre-post measures of attitudes like belonging. While these approaches are important and informative, they risk reducing inclusion to a checkbox for evaluation: something that can be verified and "set aside", rather than a lived experience that unfolds during learning activity itself.

This limitation is particularly consequential in collaborative problem-solving (CPS) contexts \cite{nixon2024catalyzing}. CPS depends not only on cognitive coordination, but also on social and relational conditions: who participates, whose contributions are acknowledged, how disagreement is handled, and whether learners feel safe contributing \cite{dowell2019promoting,choi2025agentic}. Prior work consistently demonstrates that inclusive climates and psychological safety shape participation, coordination, and team performance \cite{han2024revisiting,schneider2021collaboration}. Particularly, in STEM settings, establishing inclusive team climates remains a persistent challenge, as women and underrepresented racial minorities often report fewer opportunities to contribute, diminished recognition of their ideas, and reduced interpersonal power in group interactions. These moment-to-moment experiences accumulate over time, shaping learners’ sense of belonging and long-term engagement in STEM fields \cite{dasgupta2015female}.

The growing presence of AI in collaborative learning introduces an additional layer to this challenge. With recent advances in large language models and agentic AI systems, AI is no longer merely a computational tool. Instead, AI can participate directly in group activity as a teammate -- suggesting ideas, evaluating options, coordinating decisions, and shaping the flow of collaboration. This shift creates both opportunities and risks. AI teammates may help groups work more effectively, but they may also reshape social dynamics in ways that reinforce existing inequities, particularly for participants whose voices are already less likely to be recognized. Understanding these effects requires analytic tools that can capture inclusion as it unfolds during interaction, rather than only as a retrospective perception.

Despite the importance of these concerns, progress has been constrained by a persistent measurement gap: inclusion and sense of belonging are typically conceptualized as affective states and assessed through self-report instruments. Although valuable, these measures provide limited visibility into how inclusion unfolds moment-by-moment during collaboration—who speaks, whose ideas are taken up, how disagreement is navigated, and how affective tone shapes participation. This limitation is particularly problematic in human–AI teaming contexts, where AI participation may redistribute attention, authority, and epistemic weight in ways that are not easily captured by surveys alone.\textit{ }Ironically, this gap persists despite the AIED community’s unique capacity to model learning as a dynamic and interactional process, suggesting an opportunity---and perhaps an obligation---to develop theoretically grounded, analytic approaches that surface the mechanisms and temporal trajectories through which inclusion is produced or undermined during collaborative activity.

To address this gap, we introduce the concept of \textit{inclusion analytics}: a discourse-based measurement framework that operationalizes inclusion through observable interaction patterns. We conceptualize inclusion along three complementary dimensions, namely participation equity, affective climate, and epistemic equity -- each capturing a distinct aspect of learners’ lived experiences in collaborative activity. Using empirical data from human-AI teaming experiments, complemented by simulated collaborative discourse designed to isolate and illustrate specific inclusion mechanisms, we present an initial proof of concept demonstrating how inclusion analytics can surface inequities that remain invisible to aggregated or post-hoc measures. More broadly, we argue that inclusion analytics represents a methodological direction for the AIED community, one that leverages its theoretical foundations and technical capabilities to move beyond checklists and self-reports toward a richer, process-oriented understanding of inclusion in human-AI collaboration.

\section{Inclusion Analytics: A Multi-Dimensional Framework}\label{in-a}
In this paper, we conceptualize inclusion in CPS as an interactional construct that is enacted through observable patterns of discourse rather than solely experienced as an internal state. Specifically, we operationalize inclusion along three complementary dimensions: \textit{participation equity} (i.e., who has access to the conversational floor), \textit{affective climate} (i.e., whether interactional norms support social safety and mutual respect), and \textit{epistemic equity} (i.e., whose ideas are taken up and shape the group’s reasoning).

\subsection{Participation Equity} \label{pe-t}
Balanced participation in CPS is widely recognized as an important condition for successful collaboration. Prior work in collaborative learning consistently shows that collaborative outcomes are more productive when participation is more evenly distributed across group members, as this allows learners to more fully leverage one another's knowledge and skills \cite{woolley_evidence_2010}. For example, groups in which conversational participation is dominated by a small subset of members are less collectively intelligent than those characterized by more equal distributions of turn-taking. Accordingly, whether learners in group have comparably distributed access to the conversational floor, also known as participation equity, represents a critical dimension for examining the effectiveness of collaborative activity.

However, participation equity is more than a performance-related concern; it is also a fundamentally an inclusion issue. CPS is an interactional process through which ideas are externalized, negotiated, and integrated through discussions. When participation is imbalanced, learners who speak less have fewer opportunities to verbalize their reasonings, explain their ideas, or engage in interactive sense-making, hence limiting opportunities for learning \cite{eddy2015caution}. Moreover, persistent participation imbalances can result in some learners' contributions remaining unheard or unacknowledged, fostering experiences of exclusion and undermining a sense of belonging, particularly for students from marginalized or underrepresented identities \cite{goldsmith2024overcoming}. 

For these reasons, we include participation equity as a core dimension of inclusion analytics, conceptualizing it as an interactional property of discourse. Consistent with prior work, we operationalize participation equity using observable indicators such as turn counts and word counts, which are aggregated into group-level indices that capture the degree of participation imbalance \cite{asano_what_2024}.

\subsection{Affective Climate} \label{ac-t}
Affective climate plays an important role in CPS as it shapes whether learners feel safe to participate in group interactions. Collaborative learning is not only a cognitive process but also a social one, in which ideas are externalized, negotiated, and integrated through discourse, and disagreement must be managed interpersonally. Accordingly, the emotional and relational tone of interaction can influence how, when, and whether learners engage in collaboration. Specifically, prior work identifies psychological safety as a key condition for effective collaboration and learning \cite{edmondson_psychological_1999}. Psychological safety refers to a shared belief that the group is safe for interpersonal risk taking, such as asking questions, proposing tentative ideas, or challenging others’ views.

Affective climate contributes to psychological safety by signaling whether such interpersonal risks are welcomed or discouraged. Supportive and respectful interactions encourage continued engagement, even under uncertainty or disagreement, whereas negative affective cues can suppress participation and amplify perceived social risk \cite{dhir_social_2025}. Affective climate is therefore a fundamentally inclusion relevant dimension of CPS. Hostile or unsafe interactional climates can produce experiences of exclusion and undermine their sense of belonging, particularly for learners from marginalized communities \cite{booker2016connection}. These dynamics are especially consequential in STEM learning environments, where women and students from underrepresented racial and ethnic groups have consistently reported lower senses of belonging and higher fear of mistakes, risk-taking and vulnerability \cite{davis2023engineering}. As such, we include affective climate as the second dimension of inclusion analytics. Whereas participation equity captures who has access to the conversational floor, affective climate captures whether the interactional environment supports social safety and mutual respect. 

In this proof-of-concept paper, we operationalize affective climate through \textit{politeness}. We focus on politeness not as a proxy for positive affect per se, but as an interactional signal through which respect, deference, and relational alignment are negotiated in discourse.  While positive affect can be shown through multiple emotional and interpersonal cues, politeness offers a well-established mechanism for maintaining relationships, facilitating smooth communication, and reduce frictions in group discussions \cite{chiu_serving_2022}.

\subsection{Epistemic Equity}\label{ee-t}

Epistemic equity concerns whether learners'  contributions are treated as legitimate knowledge resources within CPS, rather than merely whether they have opportunities to speak. In collaborative discussions, influence is not always proportional to the quality of one's contribution. Engle et al. (2014) \cite{engle_toward_2014} notes that some students exert unduly greater influence than can be explained by the normative quality of their contributions. That is, authority, privilege, and access can shape what the group treats as valuable or high quality. These dynamics matter especially because influence shapes whose ideas are incorporated into the group decisions and whose contributions are recognized as legitimate knowledge in the interaction.

Research in organizational studies further demonstrates that epistemic influence can be structured by social status rather than expertise, affecting which ideas are taken up by the group and who ultimately receives credit from them \cite{ridgeway_unpacking_2004} . For instance, social role theory and expectation states theory both suggest that hegemonic beliefs about competence--- often tied to gender, race, or other social categories--- can organize power and prestige hierarchies in groups, regardless of whether those characteristics are related to the task at hand \cite{ridgeway2000creating}. As a result, individuals who are assumed to be more competent may be granted greater credibility and influence within groups, whereas others' ideas may be discounted or overlooked even when they are solid. These mechanisms shape interactions in concrete ways: whose input is solicited, whose reasoning is deemed more trustworthy, and whose proposals are taken seriously. 

These dynamics align closely with how inclusion is often experienced and articulated by learners as "being heard" in STEM learning contexts. When someone's ideas are consistently dismissed, they are excluded not only socially but also epistemically  -- in other words, they are "wronged specifically in her capacity as a knower (p.20)" \cite{fricker2007epistemic}. This form of exclusion is consequential both for individuals, who may disengage or question their belonging, and for groups, which lose access to diverse perspectives and potentially valuable ideas. These perspectives motivate epistemic equity as the last component of inclusion analytics framework: beyond equal access to participation or psychological safety, inclusive collaboration also requires equitable recognition of ideas as epistemic contributions.

\section{Operationalizing Inclusion from Discourse}

\subsection{Participation Equity Metrics}
As discussed in Section \ref{pe-t}, participation equity in collaborative discourse is commonly measured using inequality-based metrics, such as the Gini coefficient. While multiple inequality metrics exist (e.g., Gini index, Index of Concentration; see Asano et al. (2024) \cite{asano_what_2024}), we adopt the \textbf{Inequality of Participation (IP)} measure for three reasons: it is designed to adjust for small group sizes, it is straightforward to compute from simple participation counts, and it is easy to interpret where 0 indicates perfect equal participation and 1 indicates perfect imbalance. 

Formally, IP is computed as follows: 
$$IP = \frac{\frac{1}{n}\sum(E_i-O_i)}{1-\frac{1}{n}}$$
where $n$ is the number of speakers in the group, $E_i$ is the expected cumulative participation under perfect equality, and $O_i$ is the observed cumulative participation. 

Importantly, IP is a conversation (or team) level measure. Rather than capturing individual performance, it summarizes the overall degree of participation imbalance within a group, with 0 representing perfect equal participation and 1 representing maximal participation imbalance. In this paper, we measure IP both using turn counts (turns of talk) and word counts.

\subsection{Affective Climate Metrics}
In this paper, we operationalize affective climate through politeness, as discussed in Section \ref{ac-t}. Again, we do not treat politeness as a proxy for positive affect per se. Rather, we focus on politeness as an interactional signal through which respect, deference, and relational alignment are negotiated in discourse.

Building on this conceptualization, we operationalize affective climate through politeness \textit{uptake} -- that is, the extent to which politeness language styles are reciprocated and echoed by teammates in subsequent turns, capturing whether respectful interactional norms become socially reinforced during the group's ongoing collaboration. Because politeness is multidimensional and highly context-dependent, it poses challenges for computational measurement \cite{priya_computational_2024}. Given the scope of this paper, we therefore adopt a deliberately simple bag-of-words approach grounded in established politeness markers.

Specifically, we selected 11 politeness strategy dimensions adapted from Danescu-Niculescu-Mizil et al (2013) \cite{danescu-niculescu-mizil_computational_2013}, based on their positive relevance to politeness and to collaborative problem solving contexts: gratitude, apology, greeting, deference, please, indirect, counterfactual modals, indicative modals, hedging, positive lexicons, and first-person start. Using a LLM (GPT-5.2-Thinking) and three CPS datasets collected by the authors, we generated regular expressions capturing how these strategies manifest in authentic collaborative discourse (the regular expression list is available at [redacted]).

For each utterance, we count the occurrences of words or phrases associated with each category, and compute a rate by dividing the count by the total number of tokens in the utterance, resulting in 11-dimensional politeness vector for utterance $i$. We then normalize this vector and compute its cosine similarity with the politeness vectors of the next $K$ utterances within a fixed future window, excluding utterances produced by the same speaker to ensure that uptake reflects responses by others rather than self-repetition. Then we aggregate the similarities using the mean. This value serves as the politeness uptake score for utterance $i$. 

Specifically, the \textbf{politeness uptake} score of an utterance $i$ is computed as follows.

For any utterance $i$, let $c \in \mathcal{C}$ represent the set of politeness categories (e.g., gratitude, apology). The politeness vector of utterance $i$ is represented as a vector $\mathbf{p}_i\in\mathbb{R}^{|\mathcal{C}|}$, where each element is the normalized rate of that category:

\[
p_{i,c}= \frac{\text{count}(c,i)}{\text{tokens}(i)} \]
\[ \mathbf{p}_i = [p_{i,c_1},p_{i,c_2},...,p_{i,c_{11}}]
\]

Then, let $S(i)$ be the speaker of utterance $i$. We define a set of eligible future utterances $J_i$, within a window of size $K$, excluding utterances by the same speaker: $$J_i= \{j|i<j \leq i+K, S(j) \neq S(i)\}$$. Then, we use cosine similarity to measure the alignment between the politeness vectors of utterance $i$ and future utterance $j$. To account for utterances without detected politeness markers, we adopt the following convention:
\[
\mathrm{sim}(\mathbf{p}_i, \mathbf{p}_j) =
\begin{cases}
\text{undefined}, & \|\mathbf{p}_i\| = 0, \\[6pt]
0, & \|\mathbf{p}_i\| > 0 \ \text{and}\ \|\mathbf{p}_j\| = 0, \\[8pt]
\dfrac{\mathbf{p}_i \cdot \mathbf{p}_j}{\|\mathbf{p}_i\| \, \|\mathbf{p}_j\|}, &
\|\mathbf{p}_i\| > 0 \ \text{and}\ \|\mathbf{p}_j\| > 0 .
\end{cases}
\]

That is, if my current utterance $i$ does not have any politeness markers, the uptake is undefined (i.e., no polite signal to take up). If my current utterance does have politeness markers but the future utterance $j$ does not, the similarity becomes zero.

Finally, the politeness uptake score $P_i$ is computed as the mean similarity across all eligible future utterances: $$P_i = \frac{1}{|J_i|} \sum_{j \in J_i} \text{sim}(\mathbf{p}_i, \mathbf{p}_j)$$

Note that politeness uptake reflects the \textit{echoing} of politeness patterns in subsequent $K$ utterances, not the politeness of utterance $i$ itself. Therefore, it is possible that a speaker may be polite without showing high politeness uptake score if their politeness was not reciprocated. 

\subsection{Epistemic Equity Metrics}
In this proof-of-concept paper, we operationalize epistemic equity at the interaction level as \textbf{idea uptake} --- the extent to which the semantic content of a turn is echoed in subsequent turns within a short future window. This approach is motivated by prior works in computational discourse analysis that operationalizes uptake through semantic similarity and overlap across contributions \cite{dowell2019group,demszky2021measuring,suthers_exposing_2012}. Higher uptake suggests that a contribution becomes incorporated as material for shared reasoning, whereas low uptake indicates that the idea does not propagate into the group's ongoing knowledge-building process.

However, relying solely on semantic similarity poses challenges in CPS contexts. Because CPS tasks constrain participants to reason about the same problem, contributions may appear highly similar even when an idea is ignored or dismissed. Moreover, uptake does not always occur through semantic paraphrasing; participants often signal agreement through brief endorsement expressions (e.g., “yeah, that sounds great”) rather than restating the idea, which can indicate uptake despite low semantic similarity.

To address these challenges, we introduce two adjustments. First, we decompose idea uptake into \textbf{semantic-based uptake} and \textbf{endorsement-based uptake}. Semantic-based uptake captures the extent to which the content of an utterance is semantically reflected within a short temporal window following the contribution,  whereas endorsement-based uptake captures whether explicit expressions of agreement or validation occur within that window. Similar to the development of politeness lexicon, we used GPT-5.2-Thinking model using three existing CPS datasets collected by the authors to generate endorsement phrases commonly observed in collaborative discourse (the regular expression list is available at [redacted]).

Second, we adjust semantic-based uptake to account for background semantic similarity that arises from task constraints rather than genuine uptake. Because CPS utterances are often topically aligned regardless of genuine uptake, we estimate a noise baseline using random temporal sampling and subtract this baseline from observed uptake scores, allowing us to better isolate interactional uptake from topic-driven similarity.

First, the \textbf{semantic-based uptake} of an utterance $i$ is measured as follows:
Each utterance $i$ is first transformed into a semantic vector $\mathbf{h}_i \in \mathbb{R}^d$ using a pre-trained Transformer model (sentence-BERT) \cite{reimers2019sentence}. The vectors are unit-normalized such that $\|\mathbf{h}_i\|=1$, allowing the dot product to represent cosine similarity: $\text{sim}(\mathbf{h_i},\mathbf{h_j})=\mathbf{h}_i^\top\mathbf{h}_j$.

We define a window size of $K$ and a set of eligible subsequent utterances $J_i$, excluding those from the same speaker $S(i)$:$$J_i = \{j \mid i < j \leq i + K, \quad S(j) \neq S(i)\}$$.
Then, the uptake score $U_i$ is computed as the mean similarity across all eligible future utterances:
$$U_i = \frac{1}{|J_i|} \sum_{j \in J_i} \mathbf{h}_i^\top\mathbf{h}_j$$.
Then, to account for baseline topical similarity inherent in a conversation, a null score is calculated using Monte Carlo sampling from an outside pool $O_i$. The pool $O_i$ contains utterances from the same conversation that are outside the local influence of $i$ (defined by an exclusion radius $\delta$, where  $\delta \geq K$): $$O_i = \{j \mid j \notin [i - \delta, i + \delta], \quad S(j) \neq S(i)\}$$
We perform $N$ samplings -- in each sampling $n$, we draw a random sample $X_n$ of size $K$ from $O_i$ to compute the null mean :$$\mu_{null,i}=\frac{1}{N}\sum_{n=1}^N \left( \frac{1}{K}\sum_{j \in X_r} \mathbf{h}_i^\top \mathbf{h}_j\right)$$. Finally, the adjusted semantic-uptake score of an utterance $i$ is computed as follows: $$A_i = U_i - \mu_{null,i}$$

Furthermore, the \textbf{endorsement-based uptake} is computed as follows. Here, we shift from a fixed window approach to a continuous weight function. This is because unlike ideas, which may resurface throughout a discussion, endorsement signals are most salient when they occur in immediate succession. First, the presence of an endorsement is defined by a binary indicator function $M(j)$ using a predefined set of regular expressions:
$$M(j) = 
\begin{cases}
    1 & \text{if utterance $j$ matches any pattern} \\
    0 & \text{otherwise}
\end{cases}$$
To model the diminishing influence of an utterance over time, we apply an exponential decay weight, $w(d)$ where the value of an endorsement decreases by a constant decay parameter $\lambda$ for each subsequent turn $w(d) = \lambda^{d-1}$ where $d = j-i$ represents the distance in turns between the original utterance $i$ and the future turn $j$. Then, the total weighted endorsement $E_i$ is calculated as:
\[
E_i= \sum_{j=i+1}^{i+K} \mathbb{1}[S(j)\neq S(i)]\cdot M(j) \cdot w(j-i)
\]
where $\mathbb{1}_{[S(j) \neq S(i)]}$ is an indicator function that excludes self-endorsements\footnote{we used $\lambda=0.7$ to prioritize immediate feedback while still considering delayed validation. Throughout this paper, we set the window size to $K=4$ based on qualitative analysis of the data, except for endorsement-based uptake where we used $K=3$, as endorsement typically occurs within a shorter temporal window.}.

\section{Methods}
\subsection{Simulated Conversations}
We use GPT-5.2 to generate synthetic collaborative discourse data spanning the three inclusion analytics dimensions. The model simulates three undergraduate students collaborating via a chat-based platform on the NASA Moon Survival Task \cite{joshi2005experiential}, producing conversations of approximately 70-100 utterances per team. For each dimension, we specify balanced and imbalanced conditions (i.e., where speaker 3 is instructed to (1) participate less, (2) to be impolite, or (3) to introduce ideas that are not taken up by teammates). To evaluate the behavior of the proposed metrics under these controlled conditions, we summarize results separately by inclusion dimension. Participation equity is defined at the team level; accordingly, we compare 50 balanced and 50 imbalanced teams using a Mann–Whitney U test to assess whether the metric reliably distinguishes the two conditions. In contrast, politeness uptake and epistemic equity are defined at the individual level. For these measures, our goal is calibration rather than cross-team statistical inference. We therefore report individual-level scores with ±1 standard error of the mean (SEM), focusing on relative differences across speakers within teams and verifying that the metrics behave in theoretically expected ways under balanced and imbalanced conditions.

\subsection{Human-AI Teaming Dataset}
Data were collected from a human-AI teaming experiment conducted on [name], a chat-based research platform designed for human-AI teaming experiments \cite{blind}. The platform groups participants in a shared text-based chatroom with AI teammates powered by GPT-4o. The experiment includes 37 control teams (three human participants) and 37 treatment teams (two humans and one AI), with random nicknames assigned to preserve anonymity. Teams had 20 minutes to collaborate on the NASA Moon Survival Task (see 
\cite{blind} for more details on the experiment). To validate participation equity, we compare average team-level scores between treatment and control teams. For affective climate and epistemic equity, which are defined at the individual level, we focus on comparing AI and human teammates in treatment teams.

\section{Results}
\subsection{Simulated Conversations}
\subsubsection{Participation Equity}

\begin{table}[t]
\centering
\setlength{\tabcolsep}{4pt}
\renewcommand{\arraystretch}{1.05}
\label{tab:ip_sim_validation}
\begin{tabular}{lccc}
\toprule
 & Balanced & Imbalanced & MWU \\
\midrule
Turncounts& 0.09 (0.05) & 0.45 (0.04) & 0 ($p{<}.005$)\\
Wordcounts& 0.12 (0.06) & 0.54 (0.05) & 0 ($p{<}.005$)\\
\bottomrule
\end{tabular}
\caption{Participation equity (IP) on 50  simulated data (mean(sd)); MWU reports $U,p$ (two-sided). IP closer to 0 indicates perfect balance. In the imbalanced condition, Speaker 3 participated less than Speakers 1 and 2.}
\label{pe_table}
\end{table}
As shown in Table \ref{pe_table}, participation inequity was significantly lower in the balanced condition than in the imbalanced condition. In the imbalanced case, Person 1 and 2 contributed approximately 35-40 turns each (of 81.54 total), whereas Person 3 contributed about 5 turns on average. A similar pattern was observed for word counts: Person 1 and 2 produced roughly 220 words each (of 843 words total), while Person 3 contributed about 24 words only.

\subsubsection{Affective Climate}
\begin{figure}
    \centering
    \includegraphics[width=0.9\linewidth]{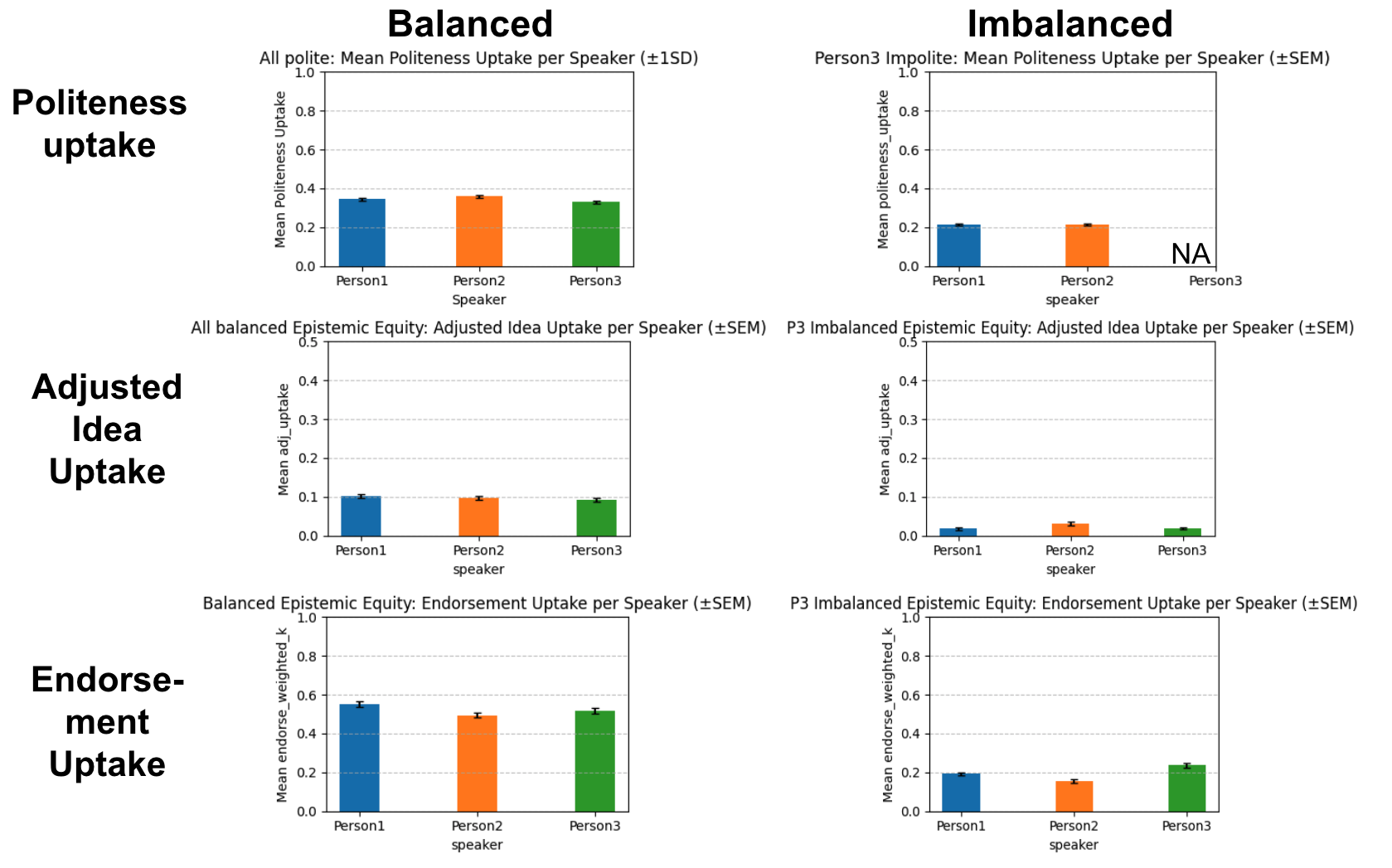}
    \caption{Average metric scores of speakers in the simulated data, with error bars showing 1 standard error of the mean. The left column shows the balanced cases, and the right column shows the imbalanced cases.}
    \label{fig:placeholder}
\end{figure}

As shown in Fig. \ref{fig:placeholder} (top), in the balanced case(left), their politeness scores were similar for all speakers, with averages around 0.35. In contrast, when only Person 3 was impolite (right), Persons 1 and 2 exhibited average politeness uptake of around 0.2, while Person 3's uptake was undefined. This follows from the definition of politeness uptake (see Section 3.2): uptake is undefined when the current utterance contains no politeness markers, and unreciprocated politeness yields zero similarity. Consequently, Person 3's uptake is undefined, whereas the presence of an impolite speaker (Person 3) reduces the average uptake for Persons 1 and 2.

\subsubsection{Epistemic Equity}
Fig. \ref{fig:placeholder} (middle, bottom) shows epistemic equity results for the conditions. Specifically, for the imbalanced condition, although Person 3 was generated such that their ideas were not taken up by teammates, both adjusted uptake and endorsement-based uptake show little separation across speakers. This occurs because, in the synthetic data, ignored ideas were often repeated or explicitly negated in subsequent turns (e.g., “Idea A doesn’t matter”), which still leads to high semantic similarity. In addition, endorsement-based uptake was inflated by generic backchanneling phrases (e.g., “yeah”) that occurred frequently in the discourse, even when ideas were not substantively adopted. 

\subsection{Human-AI Teaming Dataset}
\subsubsection{Participation Equity}
\begin{table}[t]
\centering
\setlength{\tabcolsep}{4pt}
\renewcommand{\arraystretch}{1.05}
\label{tab:hat_ip}
\begin{tabular}{lccc}
\toprule
 & Control Teams& Treatment Teams& MWU \\
\midrule
Turncounts& 0.28 (0.18)& 0.20 (0.09)& 856.5 ($p=.06$)\\
Wordcounts& 0.39 (0.17)& 0.64 (0.14)& 200 ($p{<}.005$)\\
\bottomrule
\end{tabular}
\caption{Participation equity (IP) on human-AI teaming data (mean(sd)); Mann-Whitney U reports $U,p$ (two-sided). IP closer to 0 indicates perfect balance. Treatment teams included one AI teammate and 2 humans, while control teams had 3 humans.}
\label{hat_table}
\end{table}
As shown in Table. \ref{hat_table}, control teams (three humans) and treatment teams (two humans and one AI) did not differ significantly in turn counts according to Mann-Whitney U test. However, when it comes to word counts, treatment teams exhibited significantly greater imbalance in word counts, indicating that teams with an AI teammate were more uneven. This pattern is consistent with prior work showing that AI teammates tend to be substantially more verbose than human teammates, producing nearly four times as many words on average \cite{blind}. 

\subsubsection{Affective Climate}
\textbf{\begin{table}[t]
\centering
\setlength{\tabcolsep}{4pt}
\renewcommand{\arraystretch}{1.05}
\label{tab:hat_all}
\begin{tabular}{lccc}
\toprule
 & Human & AI & MWU \\
\midrule
Politeness Uptake& 0.23 (0.22)& 0.11 (0.16)& 520158 ($p=.15$)\\
Adjusted Idea Uptake& 0.11 (0.13)& 0.05 (0.11)& 626532 ($p{<}.005$)\\
Endorsement Uptake& 0.84 (0.62)& 0.25 (0.44)& 816293 ($p{<}.005$)\\
\bottomrule
\end{tabular}
\caption{Participation equity (IP) on human-AI teaming data (mean(sd)); Mann-Whitney U reports $U,p$ (two-sided). IP closer to 0 indicates perfect balance. Treatment teams included one AI teammate and 2 humans, while control teams had 3 humans.}
\label{hat_table}
\end{table}}
As shown in Table \ref{hat_table}, when it comes to politeness uptake, there was no meaningful difference between human and AI ($p=0.15$). Although descriptively humans showed slightly higher mean uptake, this difference was not statistically significant. Prior qualitative findings \cite{blind} suggest that participants often treated the AI as an assistant --- issuing direct requests with few politeness markers --- while the AI tend to use polite language. This asymmetry may contribute to higher uptake for humans, but overall low levels of explicit politeness in short, anonymous chat-based interactions may limit observable differences.

\subsubsection{Epistemic Equity}
As shown in Table \ref{hat_table}, we found statistical significant difference between human and AI for adjusted idea uptake ($p<0.005$) and endorsement uptake ($p<0.005$). In both cases, humans showed higher uptake scores than the AI. While this may appear to contrast with prior findings that participants treated the AI as an assistant, it may reflect that AI contributions are often accepted with limited subsequent discussion, whereas human contributions more frequently elicit follow-up or endorsement. For endorsement uptake, the asymmetry is particularly expected -- while the AI frequently uses endorsement language in responses to humans, participants rarely reciprocate such endorsements toward the AI, and direct acknowledgment only toward human teammates.

\section{Discussion}
This proof-of-concept paper introduced inclusion analytics as a discourse-based framework for examining inclusion as it unfolds during CPS, instead of treating inclusion as a static or post-hoc outcome. We operationalized inclusion along three complementary dimensions ---participation equity, affective climate, and epistemic equity ---using simple, non-annotated, temporally sensitive interaction-level metrics. Our goal was not to offer definitive measures, but to demonstrate that theoretically grounded inclusion constructs can be made analytically visible using scalable, interaction-level approaches.

Across both simulated and real human-AI teaming datasets, the metrics showed mixed validation results. Participation equity was the most robust. Affective climate, operationalized through politeness uptake, behaved as expected in simulated data but showed no statistically significance differences between humans and AI in human-AI teams, highlighting challenges in measuring politeness in anonymized and relatively informal CPS settings. Epistemic equity proved the most difficult to capture: neither adjusted semantic uptake nor endorsement-based uptake clearly separated epistemically marginalized speakers in simulated data, while they did between humans and AI in human-AI teams, suggesting challenges such as task-driven semantic similarity and generic backchanneling that inflate the uptakes even when ideas are ignored.

These findings emphasize both the promise and limitations of our current version of inclusion analytics. Our approach intentionally prioritizes simplicity and transparency, relying on non-annotated data and rule-based or embedding-based methods to enable scalable analysis. However, the uneven performance of some metrics suggests that more sophisticated modeling using state-of-the-art machine learning may be necessary in the future to capture subtle forms of affective and epistemic marginalization that are not reducible to surface-level similarity or lexical level endorsement. In addition, our empirical evaluation is limited to a single real-world dataset and relies on synthetic data; further validation across diverse empirical contexts remains an important direction for future work.

We therefore position this paper as an early step rather than a finished solution.  This work represents an initial effort to measure inclusion as it unfolds during collaborative interaction, grounded in a theoretically motivated framework. We invite the AIED community to build on this proof of concept -- through richer modeling approaches, shared benchmarks, and continued theoretical refinement -- to advance how inclusion is measured, interpreted, and ultimately designed for human-AI collaborative learning systems.

\begin{acks}
This work was supported in part by the Jacobs Foundation (Grant No. 2024-1533-00). 
\end{acks}

\bibliographystyle{ACM-Reference-Format}
\bibliography{bib}

\appendix

\section{Politeness Uptake: Regular Expression List }

\begin{table}[t]
\centering
\footnotesize
\setlength{\tabcolsep}{6pt}
\renewcommand{\arraystretch}{1.15}
\begin{tabular}{>{\raggedright\arraybackslash}p{0.22\linewidth} >{\raggedright\arraybackslash}p{0.74\linewidth}}
\toprule
Category & Regex \\
\midrule

gratitude &
\makecell[l]{\ttfamily\detokenize{\b(?:}\\
\ttfamily\detokenize{thank(?:s|you|u)?|}\\
\ttfamily\detokenize{thanks\s+a\s+lot|thanks\s+so\s+much|thank\s+you\s+so\s+much|}\\
\ttfamily\detokenize{much\s+appreciated|appreciat(?:e|es|ed|ing|ion)\w*|}\\
\ttfamily\detokenize{grateful|gratitude|}\\
\ttfamily\detokenize{cheers|}\\
\ttfamily\detokenize{thx+|ty\b}\\
\ttfamily\detokenize{)\b}} \\
\midrule

apology &
\makecell[l]{\ttfamily\detokenize{\b(?:}\\
\ttfamily\detokenize{sorr+y+|sorry\s+about\s+that|}\\
\ttfamily\detokenize{my\s+bad|}\\
\ttfamily\detokenize{apolog(?:y|ies|ize|ized|ising|izing)\w*|}\\
\ttfamily\detokenize{excuse\s+me|pardon\s+me|}\\
\ttfamily\detokenize{i\s+didn'?t\s+mean\s+to|did\s+not\s+mean\s+to}\\
\ttfamily\detokenize{)\b}} \\
\midrule

greeting &
\makecell[l]{\ttfamily\detokenize{\b(?:}\\
\ttfamily\detokenize{hi+|hey+|hello+|helloo+|hellooo+|}\\
\ttfamily\detokenize{yo+|hiya+|}\\
\ttfamily\detokenize{good\s+morning|good\s+afternoon|good\s+evening|}\\
\ttfamily\detokenize{what'?s\s+up|whats\s+up}\\
\ttfamily\detokenize{)\b}} \\
\midrule

deference &
\makecell[l]{\ttfamily\detokenize{\b(?:with\s+respect|respectfully|if\s+i\s+may)\b}} \\
\midrule

indirect &
\makecell[l]{\ttfamily\detokenize{\b(?:by the way|btw)\b}} \\
\midrule

please &
\makecell[l]{\ttfamily\detokenize{\b(?:please|pls|plss+|plz)\b}} \\
\midrule

counterfactual\_modal &
\makecell[l]{\ttfamily\detokenize{\b(?:could|would|might|may)\b(?:\s+you)?\b}\\
\ttfamily\detokenize{|\bif\s+possible\b|\bif\s+you\s+can\b}} \\
\midrule

indicative\_modal &
\makecell[l]{\ttfamily\detokenize{\b(?:can|will|shall)\b(?:\s+you)?\b}} \\
\midrule

hedging &
\makecell[l]{\ttfamily\detokenize{\b(?:}\\
\ttfamily\detokenize{maybe|perhaps|possibly|probably|likely|unlikely|apparently|}\\
\ttfamily\detokenize{i\s+think|i\s+believe|i\s+feel(?:\s+like)?|i\s+guess|i\s+suppose|i\s+wonder|}\\
\ttfamily\detokenize{i'?m\s+not\s+sure|im\s+not\s+sure|not\s+sure|}\\
\ttfamily\detokenize{idk|i\s+don'?t\s+know|i\s+dont\s+know|}\\
\ttfamily\detokenize{it\s+seems(?:\s+like)?|seems(?:\s+like)?|}\\
\ttfamily\detokenize{it\s+looks\s+like|looks\s+like|}\\
\ttfamily\detokenize{it\s+appears|it\s+appears\s+that|}\\
\ttfamily\detokenize{as\s+far\s+as\s+i\s+know|from\s+what\s+i\s+understand}\\
\ttfamily\detokenize{)\b}} \\
\midrule

positive\_lexicon &
\makecell[l]{\ttfamily\detokenize{\b(?:}\\
\ttfamily\detokenize{nice|great|awesome|amazing|wonderful|excellent|fantastic}\\
\ttfamily\detokenize{|good|cool|sweet|love\s+that|wow+|yay+|}\\
\ttfamily\detokenize{good\s+idea|great\s+idea|nice\s+idea|}\\
\ttfamily\detokenize{good\s+point|great\s+point|well\s+done|nice\s+work}\\
\ttfamily\detokenize{)\b}} \\
\midrule

first\_person\_start &
\makecell[l]{\ttfamily\detokenize{^(?:}\\
\ttfamily\detokenize{i\s+(?:think|feel|guess|wonder|was\s+wondering|just\s+wanted|want|wanted)|}\\
\ttfamily\detokenize{i'?m\s+(?:thinking|wondering|not\s+sure)|}\\
\ttfamily\detokenize{im\s+(?:thinking|wondering|not\s+sure))\b}} \\

\bottomrule
\end{tabular}
\caption{Politeness categories and corresponding regex patterns.}
\label{tab:politeness_regex}
\end{table}

\section{Endorsement Uptake: Regular Expression List}

\begin{table}[t]
\centering
\footnotesize
\setlength{\tabcolsep}{6pt}
\renewcommand{\arraystretch}{1.15}
\begin{tabular}{>{\raggedright\arraybackslash}p{0.22\linewidth} >{\raggedright\arraybackslash}p{0.74\linewidth}}
\toprule
Category & Regex \\
\midrule

endorsement &
\makecell[l]{\ttfamily\detokenize{\b(}\\
\ttfamily\detokenize{i\s+agree+|agree\s+with|agree|exactly|absolutely|definitely|def|for\s+sure|}\\
\ttfamily\detokenize{you'?re\s+right|you\s+are\s+right|right|true+|facts|100\%|yea|}\\
\ttfamily\detokenize{good\s+idea|great\s+idea|awesome\s+idea|solid\s+idea|nice\s+idea|smart\s+idea|}\\
\ttfamily\detokenize{fantastic|great\s+minds|great|awesome|perfect|}\\
\ttfamily\detokenize{good\s+point|great\s+point|fair\s+point|}\\
\ttfamily\detokenize{makes\s+sense|sounds\s+good|sounds\s+great|sounds\s+right|works\s+for\s+me|}\\
\ttfamily\detokenize{that'?s\s+(?:good|great|right|smart)|}\\
\ttfamily\detokenize{let'?s\s+do\s+that|let'?s\s+go\s+with|let'?s\s+run\s+with|}\\
\ttfamily\detokenize{we\s+should\s+(?:do|go\s+with)\s+that|}\\
\ttfamily\detokenize{i'?m\s+on\s+board|im\s+on\s+board|}\\
\ttfamily\detokenize{i'?m\s+down|im\s+down|}\\
\ttfamily\detokenize{\+1|second\s+that|this}\\
\ttfamily\detokenize{)\b}} \\
\bottomrule
\end{tabular}
\caption{Endorsement regex patterns.}
\label{tab:endorse_regex}
\end{table}

\end{document}